\pdfoutput=1

\documentclass[11pt]{article}

\usepackage[]{emnlp2021}

\usepackage{times}
\usepackage{latexsym}

\usepackage[T1]{fontenc}

\usepackage[utf8]{inputenc}

\usepackage{microtype}

\usepackage{amsmath}
\usepackage{amssymb}
\usepackage{bbm}
\usepackage{graphicx}
\usepackage{array,multirow}
\usepackage{float}
\usepackage{threeparttable}
\usepackage{enumerate}
\usepackage{comment}

\title{PAUSE: Positive and Annealed Unlabeled Sentence Embedding}

\author{
  Lele Cao$^1$ \;
  Emil Larsson$^{1,2}$ \;
  Vilhelm von Ehrenheim$^1$ \\
  {\bf Dhiana Deva Cavalcanti Rocha}$^1$ \;
  {\bf Anna Martin}$^1$ \;
  {\bf Sonja Horn}$^1$ \\
 $^1$Motherbrain, EQT Group, Stockholm, Sweden \\
 $^2$Modulai, Stockholm, Sweden \\
\texttt{\{lele.cao,vilhelm.vonehrenheim\}@eqtpartners.com} \\
\texttt{\{dhiana.deva,anna.martin\}@eqtpartners.com} \\
\texttt{emil@modulai.io} \; \texttt{sonja@eqtventures.com}
}

\begin{document}
\maketitle
\begin{abstract}
Sentence embedding refers to a set of effective and versatile techniques for converting raw text into numerical vector representations that can be used in a wide range of natural language processing (NLP) applications. The majority of these techniques are either supervised or unsupervised. Compared to the unsupervised methods, the supervised ones make less assumptions about optimization objectives and usually achieve better results. However, the training requires a large amount of labeled sentence pairs, which is not available in many industrial scenarios. To that end, we propose a generic and end-to-end approach -- PAUSE (Positive and Annealed Unlabeled Sentence Embedding), capable of learning high-quality sentence embeddings from a partially labeled dataset. We experimentally show that PAUSE achieves, and sometimes surpasses, state-of-the-art results using only a small fraction of labeled sentence pairs on various benchmark tasks. When applied to a real industrial use case where labeled samples are scarce, PAUSE encourages us to extend our dataset without the liability of extensive manual annotation work.

\end{abstract}

\section{Introduction}
A sentence embedding is a numerical representation used to describe the meaning of an entire sentence. Embeddings of this type are becoming increasingly important for many downstream tasks in the language understanding domain, such as similarity or sentiment analysis. Some earlier methods, like GloVe \cite{pennington2014glove}, BERT \cite{devlin2019bert} and RoBERTa \cite{liu2019roberta} pool directly from underlying token-level embeddings to create a sentence representation. Recently, these pooling strategies have been challenged by various parameterized policies that can be optimized on domain specific tasks. The majority of these are either {\it unsupervised} or {\it supervised}. While unsupervised methods only utilize unlabeled sentences, supervised methods can quickly customize the embeddings by using domain specific labels. As a consequence, supervised methods make less assumptions about optimization objectives and usually achieve better results. However, supervised training requires a large amount of labeled sentence pairs, which is usually unavailable. In many real scenarios, the dataset turns out to be {\it positive-unlabeled} (i.e.~{\it PU dataset}), where the majority is unlabeled and the rest of the samples are labeled as positive. The methods that enable learning binary classifiers on PU datasets are called {\it PU learning}. To bridge the gap between supervised and unsupervised approaches, we incorporate state-of-the-art PU learning with the general supervised sentence embedding approaches, proposing a novel method -- PAUSE (Positive and Annealed Unlabeled Sentence Embedding)\footnote{The source code, pre-processed data, and trained models are accessible publicly from: \url{https://github.com/EQTPartners/pause}}. The main highlights of PAUSE include:
\begin{enumerate}[{(1)}]
\item good sentence embeddings can be learned from datasets with only a few positive labels;
\item it can be trained in an end-to-end fashion;
\item it can be directly applied to any dual-encoder model architecture;
\item it is extended to scenarios with an arbitrary number of classes;
\item polynomial annealing of the PU loss is proposed to stabilize the training;
\item our experiments show that PAUSE constantly outperforms baseline methods.
\end{enumerate}

\section{Related Work}
Among {\it unsupervised} sentence embedding methods, some are capable of exploring the relations among sub-sentences, such as skip-thoughts \cite{kiros2015skip}, FastSent \cite{hill2016learning}, quick-thoughts \cite{logeswaran2018efficient} and DiscSent \cite{jernite2017discourse}. These methods assume that adjacent sentences always have similar semantics. However, not every corpus is long enough, perfectly ordered or coherent enough to fulfill that assumption, which limits their applicable domains. As a result, other unsupervised methods merely focus on the internal structures within each sentence, such as paragraph-vectors \cite{le2014distributed}, Doc2VecC \cite{chen2017efficient}, Sent2Vec \cite{pagliardini2018unsupervised,gupta2019better}, WMD \cite{wu2018word}, GEM \cite{yang2019parameter} and IS-BERT \cite{zhang2020unsupervised}. In general, those unsupervised approaches optimize objectives based on assumptions, which limits their embeddings from being adapted towards different applications. Recently, several concurrently proposed methods, such as \citealt{yan-etal-2021-consert,kim-etal-2021-self,carlsson2021semantic,giorgi-etal-2021-declutr}, adopt contrastive objectives by constructing different views from the same
sentence. \citealt{gao2021simcse} achieved superior results by simply using dropout to create different views. 

The {\it supervised} approaches, on the other hand, are usually (1) trained in an end-to-end manner, (2) following a dual encoder architecture and (3) finetuned from a model pretrained on SNLI (Stanford Natural Language Inference) \cite{bowman2015large} and Multi-Genre NLI \cite{williams2018broad} datasets. NLI is the task of determining whether a {\it hypothesis} is true (entailment), false (contradiction), or undetermined (neutral) given a {\it premise}\footnote{\url{http://nlpprogress.com}}. The recent representative methods include InferSent \cite{conneau2017supervised}, USE (Universal Sentence Encoder) variants \cite{cer2018universal,chidambaram2019learning}, SBERT (Sentence-BERT) \cite{reimers2019sentence} and LaBSE (Language-agnostic BERT Sentence Embedding) \cite{feng2020language}. Built upon pretrained models, they can effectively learn good embeddings from the labeled sentence pairs. However, this approach is not feasible in scenarios where the quantity of annotations is limited.


Rather than purely labeled or unlabeled, in many real scenarios, the dataset turns out to be {\it positive-unlabeled} (PU), where a small portion of sentence pairs are labeled as positive samples and the rest are unlabeled. To address this type of problem, the gap between supervised and unsupervised methods has to be filled. \citeauthor{levi2018connecting} experimented with incorporating unsupervised regularization criteria in the supervised loss. Although \cite{levi2018connecting} reported better generalization capability, all of the samples still have to be labeled. \citeauthor{jiang2018learning} made an early attempt to apply PU learning -- particularly for matrix factorization \cite{yu2017selection} -- to obtain word embeddings for low-resource languages. 
Existing PU learning methods can be divided into three categories based on how unlabeled data is handled. The first category, with methods like \cite{li2003learning,yang2017positive}, tries to assign labels to unlabeled data in a heuristic-driven and iterative manner which makes the training scattered in steps/phases and hard to implement in practice. The second includes methods like \cite{liu2003building,lee2003learning}, which treats unlabeled data as negative with lower confidence. This can be more computationally expensive to tune. The third category, with methods such as uPU \cite{du2014analysis}, nnPU \cite{kiryo2017positive}, PUbN \cite{hsieh2019classification} and Self-PU \cite{chen2020self} regards each unlabeled sample as a weighted mixture of being positive and negative. This third category optimizes a so called PU loss, and has recently become dominating due to its wide applicability and end-to-end nature. However, a major limitation is that these algorithms are only applicable to binary classification problems. In this work, we show how to adapt PU learning to effectively learn sentence embeddings from multi-class PU datasets.

\section{The Proposed Method}

\begin{figure}
    \centering
    \includegraphics[width=0.48\textwidth]{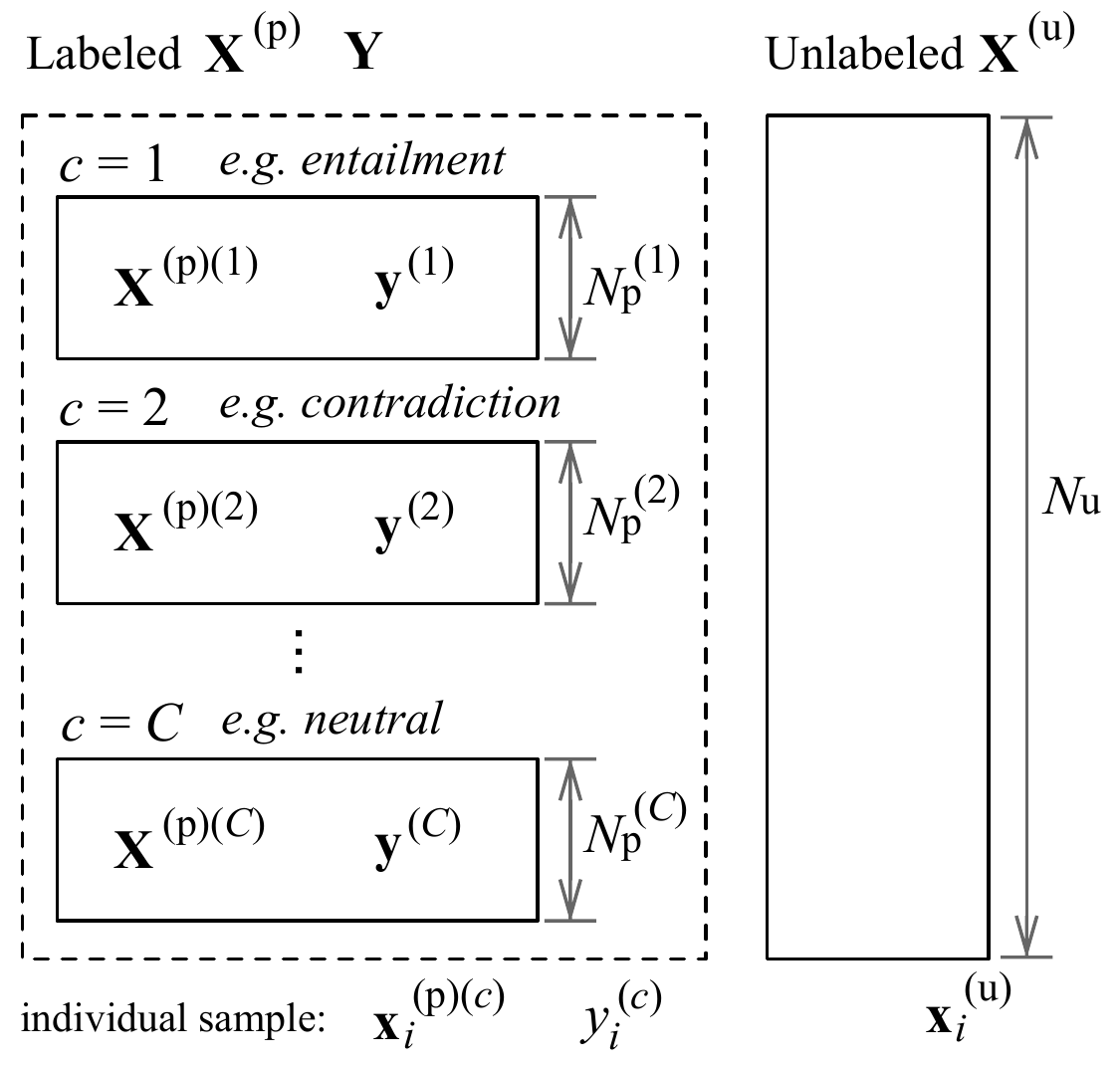}
    \caption{Problem setting and notations.}
    \label{fig:dataset}
\end{figure}

We propose a generic and end-to-end approach to obtain sentence embeddings in a setup that is a generalized natural language inference (GNLI) task. This approach can have (1) any number of classes and (2) the majority of the sentence pairs unlabeled. Let $\mathbf{X}$ be the set of sentence pairs in the entire dataset; as illustrated in Figure~\ref{fig:dataset}, $\mathbf{X}=\cup_{c=1}^{C}\mathbf{X}^{(\text{p})(c)}\cup\mathbf{X}^{(\text{u})}$, where $C(\geq2)$ is the total number of entailment classes, $\mathbf{X}^{(\text{p})(c)}$ denotes the $N_\text{p}^{(c)}$ sentence pairs labeled as the $c$-th class, and $\mathbf{X}^{(\text{u})}$ represents the $N_u$ unlabeled pairs. For the $N_\text{p} = \sum_{c=1}^C N_\text{p}^{(c)}$ labeled pairs, we use $\mathbf{Y}\in\mathbb{R}^{N_\text{p}\times C}$ to denote their mutually exclusive  and one-hot encoded labels, hence the binary label for the $c$-th entailment class should have the form of $\mathbf{y}^{(c)}\in\mathbb{R}^{N_\text{p}^{(c)}}$. On the individual sample level, we use $\mathbf{x}_i^{(\text{p})(c)}$, $y_i^{(c)}$, and $\mathbf{x}_i^{(\text{u})}$ to denote the $i$-th sentence pair that is labeled as class $c$, the binary label towards the $c$-th class for the $i$-th pair, and the $i$-th unlabeled sample respectively.

\subsection{Dual encoder model architecture}
\label{sec:method-pu-loss}
\begin{figure}
    \centering
    \includegraphics[width=0.48\textwidth]{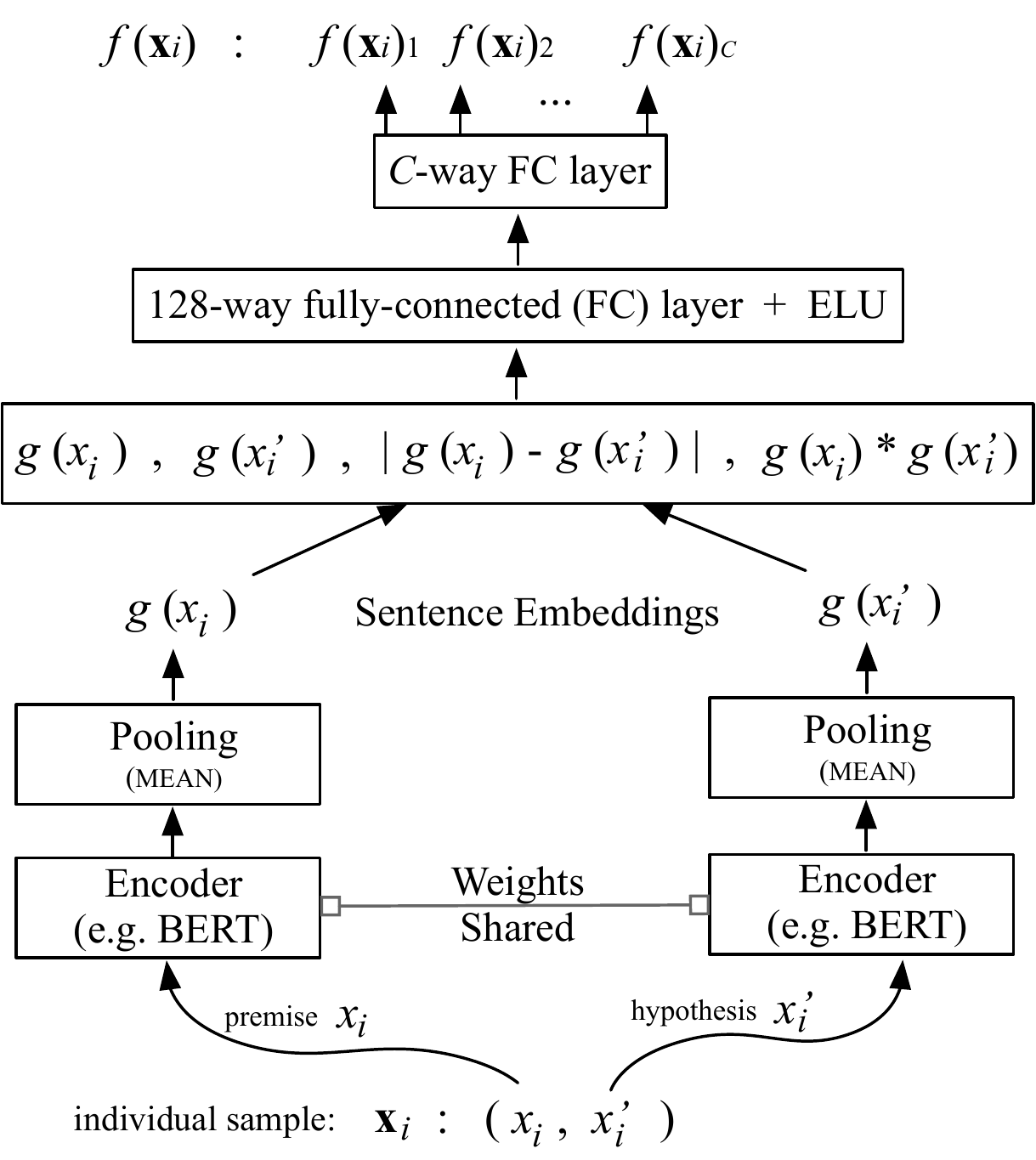}
    \caption{The dual encoder model architecture.}
    \label{fig:model}
\end{figure}

The model architecture of PAUSE follows a dual encoder schema (Figure~\ref{fig:model}) that is widely adopted in supervised sentence embedding training. Each individual sample $\mathbf{x}_i$ contains a pair of hypothesis and premise sentences $(x_i, x'_i)$, each of which is fed into a pretrained encoder (e.g. BERT). As shown in Figure~\ref{fig:model}, the two encoders are identical during the training by sharing their weights. We add a pooling operation to the output of both encoders to obtain the fixed sized sentence embeddings $g(x_i)$ and $g(x'_i)$, and following the empirical suggestion of \cite{reimers2019sentence}, we apply the \verb|MEAN|-strategy (i.e. calculating the average of the encoder output vectors). Once the sentence embeddings are generated, three matching methods are applied to extract relations between $g(x_i)$ and $g(x'_i)$: (1) concatenation of the two vectors, (2) absolute element-wise difference $|g(x_i)-g(x'_i)|$, and (3) element-wise product $g(x_i)*g(x'_i)$. The results of the three matching methods are then concatenated into a vector, which captures information from both the premise and the hypothesis. This vector is fed into a 128-way fully-connected (FC) layer with ELU (Exponential Linear Unit) activation \cite{clevert2015fast}, the output of which is transformed by a $C$-way linear FC layer, obtaining the final output $\boldsymbol{f}(\mathbf{x}_i)=[f(\mathbf{x}_i)_1, f(\mathbf{x}_i)_2, \ldots, f(\mathbf{x}_i)_C]$.

\subsection{Supervised loss}
For multi-class and mono-label problems, we calculate cross entropy (CE) loss using the labeled samples:
\begin{equation}
\label{eq:ce-loss}
\begin{aligned}
\mathcal{L}_{\text{CE}}\!=\!
  \frac{-1}{CN_{\text{p}}}
  \sum_{i=1}^{N_{\text{p}}}
  \sum_{c=1}^C
  \left|y_i^{(c)}\right|\log\left(
    \frac{e^{f(\mathbf{x}_i)_c}}
    {\sum_{j=1}^C e^{f(\mathbf{x}_i)_j}}
  \right).
\end{aligned}
\end{equation} 
For multi-class multi-label problems, the supervised loss can be binary CE:
\begin{equation}
\label{eq:bce-loss}
\begin{aligned}
\mathcal{L}_{\text{CE}}\!=\!
  \frac{-1}{CN_{\text{p}}}
  \sum_{i=1}^{N_{\text{p}}}
  \sum_{c=1}^C
  \left|y_i^{(c)}\right|\log\left(
    \frac{1}
    {1+e^{-f(\mathbf{x_i})_c}}
  \right).
\end{aligned}
\end{equation} 
When there is absolutely no negative label for binary classification problems, the supervised loss can be safely ignored. 

\subsection{Positive unlabeled loss}

To leverage unlabeled data in obtaining better sentence embeddings, 
we turn to the state-of-the-art PU learning methods, among which we largely follow \cite{du2014analysis,du2015convex,kiryo2017positive} due to their effectiveness and simplicity. Recently, \citeauthor{chen2020self} proposed an updated version which achieved marginal improvement but with the cost of greatly increased training complexity.
To facilitate computing PU loss, we address each class separately as a binary classification problem, therefore $y_i^{(c)}\in\{\pm 1\}$. For each class $C$, we define $p(\mathbf{x}, y^{(c)})$ as the joint density of $(\mathbf{X}, \mathbf{y}^{(c)})$, $p_{\text{p}}^{(c)}(\mathbf{x})=p(\mathbf{x}|y^{(c)}=+1)$ as the positive marginal, $p_{\text{n}}^{(c)}(\mathbf{x})=p(\mathbf{x}|y^{(c)}=-1)$ as the negative marginal, $p^{(c)}(\mathbf{x})$ as the unlabeled marginal, $\pi_{\text{p}}^{(c)}=p(y^{(c)}=+1)$ as the positive prior, and $\pi_{\text{n}}^{(c)}=1-\pi_{\text{p}}^{(c)}$ as the negative prior.

Assuming that we have all samples labeled for the $c$-th class, we can easily estimate the error risk $\mathcal{R}(f)_c$ as
\begin{equation}
\label{eq:r_f_c}
\begin{aligned}
\mathcal{R}(f)_c =
  \mathbb{E}_{(\mathbf{x}, y^{(c)})\sim p(\mathbf{x}, y^{(c)})} 
    [\ell(f(\mathbf{x})_c, y^{(c)})]
\end{aligned},
\end{equation}
where function $f$ is approximated by the model depicted in Figure~\ref{fig:model}, and $\ell:\mathbb{R}\times\{\pm1\}\to\mathbb{R}$ is the loss function, such that the value $\ell(a,b)$ means the loss incurred by predicting an output $a$ when the ground truth is $b$. The feasible $\ell$ functions can be referred to in \cite{kiryo2017positive}. Noticing that $\mathcal{R}(f)_c$ can be equivalently calculated by summing the error from positive and negative samples (denoted as $\mathcal{R}_{\text{p}}^{+}(f)_c$ and $\mathcal{R}_{\text{n}}^{-}(f)_c$ respectively):
\begin{equation}
\label{eq:pn-loss}
\begin{aligned}
\mathcal{R}(f)_c & =  \pi_{\text{p}}^{(c)} \mathcal{R}_{\text{p}}^{+}(f)_c 
    +  \pi_{\text{n}}^{(c)} \mathcal{R}_{\text{n}}^{-}(f)_c \\
  & =
  \pi_{\text{p}}^{(c)}
  \mathbb{E}_{\mathbf{x}\sim p_{\text{p}}^{(c)}(\mathbf{x})} 
    [\ell^+] \\
  & + 
  \pi_{\text{n}}^{(c)}
  \mathbb{E}_{\mathbf{x}\sim p_{\text{n}}^{(c)}(\mathbf{x})} 
    [\ell^-],
\end{aligned}
\end{equation}
where $\ell^-=\ell(f(\mathbf{x})_c, -1)$ and $\ell^+=\ell(f(\mathbf{x})_c, +1)$. Since $p^{(c)}(\mathbf{x})$ = $\pi_{\text{p}}^{(c)}p_{\text{p}}^{(c)}(\mathbf{x})$ + $\pi_{\text{n}}^{(c)}p_{\text{n}}^{(c)}(\mathbf{x})$, we have:
\begin{equation}
\label{eq:pu-derive-raw}
\begin{aligned}
\mathbb{E}_{\mathbf{x}\sim p^{(c)}(\mathbf{x})}[\ell^-] & =
\pi_{\text{p}}^{(c)}
\mathbb{E}_{\mathbf{x}\sim p_{\text{p}}^{(c)}(\mathbf{x})}[\ell^-] \\
& +
\pi_{\text{n}}^{(c)}
\mathbb{E}_{\mathbf{x}\sim p_{\text{n}}^{(c)}(\mathbf{x})}[\ell^-].
\end{aligned}
\end{equation}
For the sake of simplicity, we denote the terms $\mathbb{E}_{\mathbf{x}\sim p^{(c)}(\mathbf{x})}[\ell^-]$ and $\mathbb{E}_{\mathbf{x}\sim p_{\text{p}}^{(c)}(\mathbf{x})}[\ell^-]$ as $\mathcal{R}_{\text{u}}^{-}(f)_c$ and $\mathcal{R}_{\text{p}}^{-}(f)_c$ respectively. As a result, \eqref{eq:pu-derive-raw} becomes
\begin{equation}
\label{eq:pu-derive}
\begin{aligned}
\mathcal{R}_{\text{u}}^{-}(f)_c =
\pi_{\text{p}}^{(c)}
\mathcal{R}_{\text{p}}^{-}(f)_c
+
\pi_{\text{n}}^{(c)}
\mathcal{R}_{\text{n}}^{-}(f)_c.
\end{aligned}
\end{equation}
By solving \eqref{eq:pn-loss} and \eqref{eq:pu-derive}, we can eliminate the term $\mathcal{R}_{\text{n}}^{-}(f)_c$ in \eqref{eq:pn-loss}:
\begin{equation}
\label{eq:pu-loss-compact}
\begin{aligned}
\mathcal{R}(f)_c & =  
  \pi_{\text{p}}^{(c)} \mathcal{R}_{\text{p}}^{+}(f)_c \\
  & + \left[\mathcal{R}_{\text{u}}^{-}(f)_c
  - \pi_{\text{p}}^{(c)} \mathcal{R}_{\text{p}}^{-}(f)_c \right],
\end{aligned}
\end{equation}
where the two terms are called positive risk and negative risk respectively. \citeauthor{kiryo2017positive} argue that when the value of negative risk becomes less than zero, it often indicates potential overfitting. In that circumstance, we empirically choose to drop the positive risk and optimize reversely in respect to the negative risk term. Hence, in implementation, the error risk for the $c$-th class has the form of
\begin{equation}
\label{eq:nnpu-loss}
\begin{aligned}
\mathcal{R}(f)_c & =  
  \mathbbm{1}\{\mathcal{R}_{\text{u}}^{-}(f)_c
  - \pi_{\text{p}}^{(c)} \mathcal{R}_{\text{p}}^{-}(f)_c \geq 0\} \\
  &~~~~\times \pi_{\text{p}}^{(c)} \mathcal{R}_{\text{p}}^{+}(f)_c \\
  & + \max\{\mathcal{R}_{\text{u}}^{-}(f)_c
  - \pi_{\text{p}}^{(c)} \mathcal{R}_{\text{p}}^{-}(f)_c, \\
  &~~~~~\pi_{\text{p}}^{(c)} \mathcal{R}_{\text{p}}^{-}(f)_c
  - \mathcal{R}_{\text{u}}^{-}(f)_c\}.
\end{aligned}
\end{equation}
For $\ell$, we choose to use \verb|sigmoid| loss, i.e. $\ell(a,b)$ = $(1+e^{ab})^{-1}$, and we can conveniently calculate $\mathcal{R}(f)_c$ by plugging in the flowing equations:
\begin{equation}
\label{eq:nnpu-plugging}
\begin{aligned}
\mathcal{R}_{\text{p}}^{+}(f)_c & = 
  \frac{1}{N_{\text{p}}^{(c)}}\sum_{i=1}^{N_{\text{p}}^{(c)}}
  \left(1+e^{f(\mathbf{x}_i^{(\text{p})(c)})_c}\right)^{-1},\\
\mathcal{R}_{\text{p}}^{-}(f)_c & = 
  \frac{1}{N_{\text{p}}^{(c)}}\sum_{i=1}^{N_{\text{p}}^{(c)}}
  \left(1+e^{-f(\mathbf{x}_i^{(\text{p})(c)})_c}\right)^{-1},\\
\mathcal{R}_{\text{u}}^{-}(f)_c & = 
  \frac{1}{N_{\text{u}}}\sum_{i=1}^{N_{\text{u}}}
  \left(1+e^{-f(\mathbf{x}_i^{(\text{u})})_c}\right)^{-1}.
\end{aligned}
\end{equation}
The overall PU loss $\mathcal{L}_{\text{PU}}$ can be constructed using \eqref{eq:nnpu-loss} and \eqref{eq:nnpu-plugging}:
\begin{equation}
\label{eq:pu-loss-final}
\mathcal{L}_{\text{PU}} = \frac{1}{C}\sum_{c=1}^{C}\mathcal{R}(f)_c
\end{equation}

\subsection{Annealed joint optimization}
During training, the model is optimized in an end-to-end manner on mini-batches. Every mini-batch is sampled from all subsets of the dataset (cf.~Figure~\ref{fig:dataset}) according to the relative subset sizes, so that the mini-batches reflect the composition of the entire dataset. Because the initial estimations of positive/negative risks in \eqref{eq:pu-loss-compact} tend to be inaccurate, simply optimizing both losses (i.e. $\mathcal{L}_{\text{CE}}$ and $\mathcal{L}_{\text{PU}}$) jointly with the same weights often leads to sub-optimal and unstable solution. This problem is particularly prominent when the dataset is large or the model is highly flexible. As a result, we propose to apply an annealing strategy to the PU loss component when constructing the overall loss:
\begin{equation}
\label{eq:overall-loss}
\mathcal{L} = \mathcal{L}_{\text{CE}} 
+ \left(\frac{t}{T}\right)^\alpha\mathcal{L}_{\text{PU}},
\end{equation}
where $T$ denotes the total number of training steps and $1\leq t\leq T$ is the elapsed number of steps. The hyper-parameter $\alpha\geq2$ controls the annealing speed. We empirically discover that $\alpha=3$ usually offers optimal and stable performance. For binary classification problems ($C=1$), the overall loss will fall back to $\mathcal{L}_{\text{PU}}$ when there is no negative labels available.

\begin{table*}[h]
\addtolength{\tabcolsep}{-1.2pt}
\centering
\begin{threeparttable}
\begin{tabular}{c|l|ccccccc|c}
\hline
\multicolumn{2}{c|}{\textbf{Model}} & \textbf{STS12} & \textbf{STS13} & \textbf{STS14} & \textbf{STS15} & \textbf{STS16} & \textbf{STSb} & \textbf{SICK-R} & \textbf{Avg.} \\
\hline
\parbox[t]{2mm}{\multirow{5}{*}{\rotatebox[origin=c]{90}{unsupervised}}} &
FastSent\tnote{*} & - & - & 63.00 & - & - & - & 61.00 & - \\
& Avg. GloVe embeddings\tnote{$\dagger$} & 55.14 & 70.66 & 59.73 & 68.25 & 63.66 & 58.02 & 53.76 & 61.32 \\
& Avg. BERT embeddings\tnote{$\dagger$} & 38.78 & 57.98 & 57.98 & 63.15 & 61.06 & 46.35 & 58.40 & 54.81 \\
& BERT [\verb|CLS|]-vector\tnote{$\dagger$} & 20.16 & 30.01 & 20.09 & 36.88 & 38.08 & 16.50 & 42.63 & 29.19 \\
& IS-BERT-NLI\tnote{$\ddagger$} & 56.77 & 69.24 & 61.21 & 75.23 & 70.16 & 69.21 & 64.25 & 66.58 \\
\hline
\parbox[t]{2mm}{\multirow{4}{*}{\rotatebox[origin=c]{90}{supervised}}} &
InferSent-Glove\tnote{$\dagger$} & 52.86 & 66.75 & 62.15 & 72.77 & 66.87 & 68.03 & 65.65 & 65.01 \\
& USE\tnote{$\dagger$} & 64.49 & 67.80 & 64.61 & 76.83 & 73.18 & 74.92 & {\bf 76.69} & 71.22 \\
& SBERT-NLI-base\tnote{$\dagger$} & 70.97 & {\bf 76.53} & {\bf 73.19} & {\bf 79.09} & 74.30 & 77.03 & 72.91 & 74.89 \\
& SRoBERTa-NLI-base\tnote{$\dagger$} & 71.54 & 72.49 & 70.80 & {\bf 78.74} & 73.69 & 77.77 & 74.46 & 74.21 \\
\hline
\parbox[t]{2mm}{\multirow{14}{*}{\rotatebox[origin=c]{90}{PAUSE variants}}} &
PAUSE-NLI-small-100\% & 67.40 & 68.47 & 62.34 & 67.73 & 69.08 & 71.24 & 71.65 & 68.27 \\
& PAUSE-NLI-small-70\% & 65.58 & 67.60 & 61.05 & 67.25 & 67.85 & 70.42 & 70.85 & 67.23 \\
& PAUSE-NLI-small-50\% & 65.60 & 66.95 & 61.52 & 66.87 & 66.48 & 69.35 & 70.83 & 66.80 \\
& PAUSE-NLI-small-30\% & 64.50 & 65.91 & 60.42 & 67.28 & 66.42 & 69.33 & 70.00 & 66.27 \\
& PAUSE-NLI-small-10\% & 60.85 & 65.47 & 60.90 & 66.56 & 65.99 & 68.12 & 67.24 & 65.02 \\
& PAUSE-NLI-small-5\% & 59.21 & 65.32 & 59.56 & 66.14 & 65.93 & 67.58 & 65.81 & 64.22 \\
& PAUSE-NLI-small-1\% & 57.06 & 64.00 & 59.47 & 64.28 & 63.25 & 63.14 & 61.09 & 61.76 \\
\cline{2-10}
& PAUSE-NLI-base-100\% & {\bf 73.68} & {\bf 77.08} & {\bf 74.10} & {\bf 78.58} & {\bf 76.58} & {\bf 80.34} & {\bf 74.69} & {\bf 76.44} \\
& PAUSE-NLI-base-70\% & {\bf 73.96} & {\bf 76.26} & 73.13 & 77.57 & {\bf 75.74} & {\bf 78.98} & 74.53 & {\bf 75.74} \\
& PAUSE-NLI-base-50\% & {\bf 73.72} & 75.89 & 72.80 & 77.29 & {\bf 75.40} & {\bf 78.62} & {\bf 74.55} & {\bf 75.47} \\
& PAUSE-NLI-base-30\% & 73.54 & 75.15 & 72.44 & 77.09 & 74.43 & 78.39 & 73.55 & 74.94 \\
& PAUSE-NLI-base-10\% & 71.51 & 74.46 & {\bf 73.23} & 77.16 & 74.67 & 78.22 & 73.49 & 74.68 \\
& PAUSE-NLI-base-5\% & 67.59 & 73.76 & 68.68 & 73.24 & 73.47 & 75.77 & 72.28 & 72.11 \\
& PAUSE-NLI-base-1\% & 62.44 & 72.01 & 63.88 & 69.39 & 66.32 & 70.22 & 69.64 & 67.70 \\
\hline
\end{tabular}
\begin{tablenotes}
        \scriptsize
        \item[*] The results are extracted from \cite{hill2016learning}.
        \item[$\dagger$] The results are extracted from \cite{reimers2019sentence}.
        \item[$\ddagger$] The results are extracted from \cite{zhang2020unsupervised}.
\end{tablenotes}
\end{threeparttable}
\caption{\label{table:unsupervised}
Unsupervised evaluation results: Spearman's rank correlation $\rho$ between the cosine similarity of sentence embeddings and the labels for STS tasks. $\rho\times100$ is reported here. The PAUSE results are averaged from three runs with random seeds. Bold font indicates the top-3 results on each dataset.
}
\end{table*}

\section{Experiments}

Inspired by the previous works \cite{hill2016learning,reimers2019sentence,zhang2020unsupervised}, we evaluate PAUSE on STS (Semantic Textual Similarity) and SentEval\footnote{\url{https://github.com/facebookresearch/SentEval}} in both supervised and unsupervised settings. We also show the robustness of PAUSE in a real industrial use case. In our experiments, we test two versions of PAUSE: PAUSE-base (110M parameters) and PAUSE-small (4.4M parameters), which use uncased BERT-base \cite{devlin2019bert} and BERT-small \cite{turc2019well} \footnote{\url{https://github.com/google-research/bert}} as their encoder model, respectively. PAUSE is trained by minimizing \eqref{eq:overall-loss} with $\alpha=3$ (searched 3, 4, and 5) using the Adam optimizer with learning rate 7.5e-5 (searched \{1e-3, 1e-4, 7.5e-5, 5e-5, 2.5e-5, 1e-5 and 1e-6\}). The experiments are carried out on a machine (managed virtual machine instance\footnote{\url{https://cloud.google.com/ai-platform/docs/technical-overview\#notebooks}}) with four virtual CPUs (Intel Xeon 2.30GHz), 15GB RAM, and four GPUs (NVIDIA Tesla P100).
Since PAUSE requires a large batch size to ensure enough labeled samples from each class in every mini-batch, we use a batch size of 128 and 1,024 for PAUSE-base and PAUSE-small respectively, fully utilizing the GPU capacity.

\subsection{Unsupervised STS}
\label{sec:unsupervised_sts}

\begin{table}[h]
\addtolength{\tabcolsep}{0pt}
\centering
\begin{tabular}{l|l}
\hline
\textbf{Model} & $\pmb{\rho\times100}$ \\
\hline
\multicolumn{2}{l}{{\it Not trained on STSb}}\\
\hline
Avg. BERT embeddings & $46.35$ \\ 
SBERT-NLI-base & 77.03 \\ 
SRoBERTa-NLI-base & {\bf 77.77} \\ 
\hline
\multicolumn{2}{l}{{\it Only trained on STSb}}\\
\hline
IS-BERT-STSb (ft) & 74.25 $\pm$ 0.94 \\ 
IS-BERT-STSb (ssl+ft) & 84.84 $\pm$ 0.43 \\ 
SBERT-STSb-base & 84.67 $\pm$ 0.19 \\
SRoBERTa-STSb-base & {\bf 84.92} $\pm$ 0.34 \\ 
\hline
\multicolumn{2}{l}{{\it Trained on NLI, then on STSb}}\\
\hline
SBERT-NLI-STSb-base & {\bf 85.35} $\pm$ 0.17 \\
SRoBERTa-NLI-STSb-base & 84.79 $\pm$ 0.38 \\
PAUSE-NLI-STSb-base-100\% & 84.83 $\pm$ 0.30 \\
\hline
\multicolumn{2}{l}{{\it Trained on partially labeled NLI, then on STSb}}\\
\hline
PAUSE-NLI-STSb-base-70\% & {\bf 84.02} $\pm$ 0.29 \\
PAUSE-NLI-STSb-base-50\% & 83.86 $\pm$ 0.35 \\
PAUSE-NLI-STSb-base-30\% & 83.53 $\pm$ 0.37 \\
PAUSE-NLI-STSb-base-10\% & 83.39 $\pm$ 0.44 \\
PAUSE-NLI-STSb-base-5\% & 81.77 $\pm$ 0.41 \\
PAUSE-NLI-STSb-base-1\% & 73.04 $\pm$ 0.58 \\
\hline
\end{tabular}
\caption{The supervised evaluation results (Spearman’s rank correlation) on the STS benchmark (STSb) test set. The results of the non-PAUSE baselines are extracted from \cite{reimers2019sentence} and \cite{zhang2020unsupervised}.}
\label{tab:supervised}
\end{table}

We first evaluate PAUSE on STS tasks without using any STS data for training. Specifically, we choose the datasets of STS 2012-2016 \cite{agirre2012semeval,agirre2013sem,agirre2014semeval,agirre2015semeval,agirre2016semeval}, STS benchmark (STSb) \cite{cer2017semeval}, and SICK-Relatedness (SICK-R) \cite{marelli2014sick}. These datasets have labels between 0 and 5 indicating the semantic relatedness of sentence pairs. We compare PAUSE with two groups of baselines. The first group is the unsupervised methods, which includes FastSent \cite{hill2016learning}, IS-BERT-NLI \cite{zhang2020unsupervised}, the average of GloVe embeddings, the average of the last layer representations of BERT, and the [\verb|CLS|] embedding of BERT. The second group consists of supervised approaches: InferSent-Glove \cite{conneau2017supervised}, USE \cite{cer2018universal}, SBERT, and SRoBERTa \cite{reimers2019sentence}. All models are trained on the combination of the SNLI \cite{bowman2015large} and Multi-Genre NLI \cite{williams2018broad} datasets, which contains one million sentence pairs annotated with three labels ($C=3$): {\it entailment}, {\it contradiction} and {\it neutral}. PAUSE is trained for 2 epochs with a linear learning rate warm-up over the first 10\% of the training steps.

As suggested in \cite{reimers2016task,reimers2019sentence,zhang2020unsupervised}, we calculate the Spearman's rank correlation between the cosine-similarity of the sentence embeddings and the labels, which is presented in Table~\ref{table:unsupervised}. The results show that most of the supervised methods achieve superior performances compared to unsupervised ones, which has been previously evidenced by \cite{hill2016learning,cer2018universal,zhang2020unsupervised}. PAUSE using the BERT-base encoder (PAUSE-NLI-base) performs much better than the versions using the BERT-small encoder (PAUSE-NLI-small). PAUSE-NLI-base takes on average 220 minutes to complete one epoch of training, while PAUSE-NLI-small takes merely 9 minutes. The post-fix (i.e. 1\%, 5\%, ..., 100\%) in the names of the PAUSE variants indicate the percentage of NLI labels that are used during the training. Although the performance monotonically drops when using less labels, this drop remains marginal even when only 50\% of the labels are used. PAUSE-NLI-base-100\% obtained slightly better results than SBERT-NLI-base. PAUSE-NLI-base-100\% and SBERT-NLI-base are trained on the same amount of labeled samples, yet the former achieves slightly better results, probably due to differences in (1) the BERT-base versions and/or (2) the layers following the encoding step.
Observing the average results of each PAUSE-NLI-base variant, the model trained on merely 10\% of the labels results in a performance about 2\% lower than the one relying on all labels. This demonstrates that PAUSE is a label-efficient sentence embedding approach applicable to situations where only a small number of samples are labeled.

In an attempt to train the SBERT-NLI-base model using only 1\%, 5\%, 10\%, and 30\% of the labels for 2 epochs, we found that all trials suffered from over-fitting in varying degrees. While this problem could be addressed by hyper-parameter optimization and regularization, such alterations would compromise the fairness of the comparison.

\subsection{Supervised STS benchmark}
Similar to \cite{reimers2019sentence,zhang2020unsupervised}, we use the STS benchmark (STSb) dataset \cite{cer2017semeval} to evaluate the models' performance on the supervised STS task. STSb includes 8,628 sentence pairs from the categories of {\it captions}, {\it news}, and {\it forums}. The dataset is split into train (5,749), dev (1,500), and test (1,379) subsets. We use the training set to finetune PAUSE (pretrained using partially labeled NLI) using a regression objective function. On the test set, we compute the cosine similarity between each pair of sentences. 
Since PAUSE obtains better results using BERT-base compared to BERT-small (cf. Table~\ref{table:unsupervised}), we only report the results for PAUSE-NLI-STSb-base models, which are trained with five random seeds and four epochs.

In Table~\ref{tab:supervised}, we compare PAUSE to three categories of baselines: (1) not trained on STSb at all, (2) only trained on STSb, and (3) first trained on the fully labeled NLI, then finetuned on STSb. It is clear that finetuning on STSb greatly improves the model performance and pretraining on NLI further uplifts the performance slightly. Using merely 10\% to 70\% of NLI labels, PAUSE manages to achieve results comparable to the baselines that use all NLI labels. Another interesting finding is that when pretraining PAUSE on less than 5\% of the labels, the performance becomes inferior to directly finetuning on STSb. This suggests that pretraining PAUSE with too few labeled samples may result in embeddings that are hard to finetune in downstream regression tasks. In addition, we observe a clear trend in Table~\ref{tab:supervised}: the standard deviation increases when less labels are used, which is also observed in the unsupervised experiments.

\begin{table*}
\centering
\begin{threeparttable}
\begin{tabular}{l|ccccccc|c}
\hline
\textbf{Model} & \textbf{MR} & \textbf{CR} & \textbf{SUBJ} & \textbf{MPQA} & \textbf{SST} & \textbf{TREC} & \textbf{MRPC} & \textbf{Avg.} \\
\hline
Unigram-TFIDF\tnote{*} & 73.7 & 79.2 & 90.3 & 82.4 & - & 85.0 & 73.6 & - \\
SDAE\tnote{*} & 74.6 & 78.0 & 90.8 & 86.9 & - & 78.4 & 73.7 & - \\
ParagraphVec DBOW\tnote{*} & 60.2 & 66.9 & 76.3 & 70.7 & - & 59.4 & 72.9 & - \\
FastSent\tnote{*} & 70.8 & 78.4 & 88.7 & 80.6 & - & 76.8 & 72.2 & - \\
SkipThought\tnote{*} & 76.5 & 80.1 & 93.6 & 87.1 & 82.0 & {\bf 92.2} & 73.0 & 83.50 \\
Avg. GloVe embeddings\tnote{$\dagger$} & 77.25 & 78.30 & 91.17 & 87.85 & 80.18 & 83.0 & 72.87 & 81.52 \\
Avg. BERT embeddings\tnote{$\dagger$} & 78.66 & 86.25 & {\bf 94.37} & 88.66 & 84.40 & {\bf 92.8} & 69.54 & 84.94 \\
BERT [\verb|CLS|]-vector\tnote{$\dagger$} & 78.68 & 84.85 & 94.21 & 88.23 & 84.13 & 91.4 & 71.13 & 84.66 \\
IS-BERT-task\tnote{$\ddagger$} & 81.09 & 87.18 & {\bf 94.96} & 88.75 & 85.96 & 88.64 & 74.24 & {\bf 85.91} \\
InferSent-Glove\tnote{$\dagger$} & {\bf 81.57} & 86.54 & 92.50 & {\bf 90.38} & 84.18 & 88.2 & 75.77 & 85.59 \\
USE\tnote{$\dagger$} & 80.09 & 85.19 & 93.98 & 86.70 & 86.38 & {\bf 93.2} & 70.14 & 85.10 \\
SBERT-NLI-base\tnote{$\dagger$} & {\bf 83.64} & {\bf 89.43} & {\bf 94.39} & {\bf 89.86} & {\bf 88.96} & 89.6 & {\bf 76.00} & {\bf 87.41} \\
\hline
PAUSE-NLI-base-100\% & {\bf 81.40} & {\bf 87.87} & 93.64 & 88.85 & {\bf 88.41} & 84.4 & {\bf 75.83} & {\bf 85.77} \\
PAUSE-NLI-base-70\% & 80.04 & 87.74 & 92.65 & {\bf 89.42} & {\bf 87.70} & 84.2 & 75.71 & 85.35 \\
PAUSE-NLI-base-50\% & 80.27 & {\bf 87.52} & 93.09 & 89.14 & {\bf 87.70} & 81.6 & 74.61 & 84.85 \\
PAUSE-NLI-base-30\% & 80.42 & 87.31 & 92.99 & 89.04 & 87.26 & 84.8 & 74.09 & 85.13 \\
PAUSE-NLI-base-10\% & 79.93 & 86.49 & 93.43 & 88.60 & 86.44 & 86.2 & 73.68 & 84.97 \\
PAUSE-NLI-base-5\% & 80.06 & 87.05 & 93.37 & 88.86 & 85.61 & 86.0 & 75.01 & 85.14 \\
PAUSE-NLI-base-1\% & 79.99 & 85.83 & 93.33 & 88.16 & 86.66 & 86.4 & {\bf 77.10} & 85.35 \\
\hline
\end{tabular}
\begin{tablenotes}
        \scriptsize
        \item[*] The results are extracted from \cite{hill2016learning}.
        \item[$\dagger$] The results are extracted from \cite{reimers2019sentence}.
        \item[$\ddagger$] The results are extracted from \cite{zhang2020unsupervised}; IS-BERT-task is fintuned on each of the task-specific dataset (without label) to produce sentence embeddings, which are then used for training downstream classifiers.
\end{tablenotes}
\end{threeparttable}
\caption{\label{table:seneval}
Accuracy on test fold (10 fold cross validation) using domain specific SentEval tasks. Sentence embeddings from different models are used as features to train a logistic regression classifier.
}
\end{table*}

\subsection{SentEval: domain specific tasks}
In order to give an impression of the quality of our sentence embeddings for
various domain specific tasks, we choose to evaluate PAUSE on seven SentEval tasks \cite{conneau2018senteval}: (1) {\bf TREC} - fine grained question type classification \cite{li2002learning}, (2) {\bf CR} - sentiment prediction of customer product reviews \cite{hu2004mining}, (3) {\bf MRPC} - Microsoft Research Paraphrase Corpus from parallel news sources \cite{dolan2004unsupervised}, (4) {\bf SUBJ} - Subjectivity prediction of sentences from movie reviews and plot summaries \cite{pang2004sentimental}, (5) {\bf MR} - sentiment prediction for movie reviews \cite{pang2005seeing}, (6) {\bf MPQA} - opinion polarity classification \cite{wiebe2005annotating} and (7) {\bf SST} - binary sentiment analysis \cite{socher2013recursive}. Unlike \cite{devlin2019bert} and \cite{zhang2020unsupervised} that finetune the encoder on these tasks, we directly use the sentence embeddings from PAUSE-NLI-base models (cf. Section~\ref{sec:unsupervised_sts}) as features for a logistic regression classifier that is trained in a 10-fold cross-validation setup, where the prediction accuracy is computed for the test fold.

The results can be found in Table~\ref{table:seneval}, where the top-3 results on each task are presented in bold face. Largely speaking, the sentence embeddings from SBERT and PAUSE successfully capture domain specific information with the exception of the TREC task where pretraining on question-answering data (e.g. USE) seems to be beneficial. In Table~\ref{table:unsupervised}, we have observed poor results from Avg. BERT embeddings, BERT [\verb|CLS|]-vector, PAUSE-NLI-base-5\% and PAUSE-NLI-base-1\%. However, on the selected SentEval tasks, they all achieve decent results and the performance of PAUSE does not even degrade as we use significantly less labels. This inconsistency can be explained by how we measure the model performance: on STS datasets we calculated the cosine-similarity between sentence embeddings, which treats all dimensions indifferently, while SentEval fits a logistic regression classifier to the embeddings allowing different dimensions to have different impact on the classifier's output. As a result, cosine-similarity can only be relied on when the sentence embeddings are finetuned on related datasets with a large number of labeled samples. When the sentence embeddings are directly used as input features for training discriminative models on downstream tasks, finetuning on NLI only gives approximately 1$\sim$2\% performance uplift. Moreover, PAUSE appears to be unaffected by a drastic decrease in labeled samples, which is consistent with the results of IS-BERT-task.

We also notice that unsupervised PAUSE-NLI-base-100\% performs better than SBERT-NLI-base in Table~\ref{table:unsupervised}, yet this is not the case for supervised fine-tuning (Table~\ref{tab:supervised} and \ref{table:seneval}). This might be a consequence of several differences:
(1) PAUSE uses a newer version of the pretrained BERT-base model
\footnote{\url{https://tfhub.dev/tensorflow/bert_en_uncased_L-12_H-768_A-12/3}}
compared to \cite{reimers2019sentence}, 
(2) PAUSE has an extra term $g(x_i)*g(x'_i)$ when extracting relations between two sentences, as seen in Figure~\ref{fig:model}, and (3) PAUSE treats the NLI sentence pairs with ambiguous/conflicting labels as unlabeled samples that are utilized by the PU loss term during model optimization.


\begin{table*}[h]
\footnotesize\addtolength{\tabcolsep}{0pt}\renewcommand{\arraystretch}{1.1}
\centering
\begin{tabular}{| p{0.015\linewidth} | p{0.42\linewidth} | p{0.42\linewidth} | p{0.035\linewidth}|}
\hline
{\bf ID} & {\bf Anchor Company} & {\bf Candidate Company} & {\bf label} \\
\hline
1 & Owner and operator of data centers in UK ... distribute data in data centers and the global digital economy.  & Independent co-location / data center provider in Slovenia. & 1 \\ 
\hline
2 & Provider of human-computer interaction technology designed to ... documentation is common to both CX and ES. & Developer of a mobile stock-trading application designed to make ... enabling traders to discover and invest in markets without a hassle. & 0 \\
\hline
3 & Provide a reliable and fast veterinary diagnostic service ... We intend to be your partner in the daily medical diagnosis! & Provider of laboratory services. We care about ... Biochemical and haematological examinations are available 24 hours a day. & 1 \\
\hline
4 & Manufacturer of smart electric scooters designed to offer ... noise-free scooters that run on electricity. &  A peer-to-peer rental marketplace which allows people to rent spare items ... can be borrowed within minutes. & 0 \\
\hline
\multicolumn{4}{|l|}{... ... 166,832 rows (company pairs) in total} \\
\hline
\end{tabular}
\caption{The first four samples (shortened and anonymized) of the similar company (SC) dataset used to train company embeddings for identifying similar companies. Label=0 and 1 represents dissimilar and similar respectively.}
\label{tab:dataset_similar_org}
\end{table*}

\subsection{Use case: finding similar companies}
In this section, we will discuss the potential of PAUSE on a real industrial use case from EQT Group\footnote{EQT Group is a global investment organization (\url{https://eqtgroup.com}).}. The EQT investment professionals use Motherbrain\footnote{Motherbrain is EQT's proprietary investment platform driven by diversified big data and cutting-edge algorithms. (\url{https://eqtgroup.com/motherbrain}, \url{https://eqtventures.com/motherbrain})} to accomplish tasks such as deal sourcing, market analysis and metrics benchmarking to name a few. One concrete use case is competitor mapping, where the Motherbrain users track competitors for companies they are studying (a.k.a. anchor companies). Competitors are defined as the companies running a business similar to that of the anchor company. In our database, there are more than eight million companies with textual descriptions and only 12,326 of these have annotations indicating their relatedness to other companies, forming 166,832 company pairs with noisy binary labels indicating whether they are similar (55,139 pairs) or dissimilar (111,693 pairs). Table~\ref{tab:dataset_similar_org} demonstrates our similar company (SC) dataset containing these annotated company pairs, which is divided into train (156,925), dev (4,922) and test (4,985) sets. A BERT-base encoder is finetuned using the PAUSE approach on the training set. The finetuned encoder model is served via Tensorflow Serving\footnote{\url{https://github.com/tensorflow/serving}}, which is called by an Apache Beam\footnote{\url{https://beam.apache.org}} job encoding the incoming company descriptions into company embeddings in a streaming fashion. The company embeddings are then indexed by Elasticsearch\footnote{\url{https://www.elastic.co}} to support fast similarity search initiated by platform users.

To benchmark PAUSE in this setting, we trained models using different percentages of labeled samples (100\%, 50\%, 10\%, 5\%). Table~\ref{tab:result-similar-company} shows that it is sufficient to have 10\% of the samples labeled and still reach high accuracy, precision, and recall. When all samples are labeled, the accuracy is only increased by around 3\%. In practice, these results encourage us to extend our dataset without the burden of manually labeling all samples. We also speculate that increasing the size of the dataset while maintaining the balance between labeled and unlabeled samples could improve performance further. Essentially, this implies that we can achieve results close to that of a fully labeled dataset with a fraction of the manual annotation work.

\begin{table}
\addtolength{\tabcolsep}{-0.5pt}
\centering
\begin{tabular}{l|c|c|c}
\hline
{\bf Model} & {\bf Acc.} & {\bf Precision} & {\bf Recall} \\
\hline
PAUSE-SC-100\% & 76.13 & 77.55 & 61.87 \\ 
PAUSE-SC-50\% & 75.27 & 74.30 & 64.41 \\ 
PAUSE-SC-10\% & 73.35 & 73.29 & 60.58  \\ 
PAUSE-SC-5\% & 64.40 & 63.80 & 54.07 \\ 
\hline
\end{tabular}
\caption{The performance of PAUSE (using the BERT-base encoder) trained on the similar company (SC) dataset using different percentages of labeled samples.}
\label{tab:result-similar-company}
\end{table}

\section{Conclusions and Future Work}
\label{sec:conclusion}
In this work, we attempt to bridge the gap between supervised and unsupervised sentence embedding techniques, proposing PAUSE -- a generic and end-to-end sentence embedding approach that exploits the labels and explores the unlabeled sentence pairs simultaneously. PAUSE trained on NLI datasets achieves state-of-the-art results on unsupervised STS tasks, and also performs well on many downstream domain-specific tasks. In all of our experiments, we observe that PAUSE keeps performing well with a reduced number of labeled samples, as long as more than 5-10\% of the dataset is labeled. This indicates that PAUSE is a label-efficient sentence embedding approach that can be effectively applied to datasets where only a small part is labeled while the rest remains unlabeled. We also demonstrate that PAUSE helps lower the labeling requirement for an industrial use case aimed at encoding company descriptions. In that sense, PAUSE pushes the application boundary of sentence embeddings to include many more real-world scenarios where labeled samples are scarce. The possible extensions of this work include (1) augmenting the the labels with dropout, (2) experimenting with contrastive supervised loss, and (3) exploring how PAUSE can be extended with contextual sentence embeddings.

\section*{Acknowledgements}
\href{https://eqtgroup.com/}{EQT Group} and the \href{https://eqtgroup.com/motherbrain}{Motherbrain} team have provided great support along the journey of accomplishing this work; 
particularly, we would like to appreciate the insights/support of all sorts from (alphabetically ordered) 
Alex Patow,
Andjela Kusmuk,
Andreas Beccau,
Andrey Melentyev,
Anton Andersson Andrejic,
Anton Ask \r{A}str\"{o}m
Daniel Str\"{o}m,
Daniel Wroblewski,
Elin B\"{a}cklund,
Emil Broman,
Emma Sj\"{o}str\"{o}m,
Erik Ferm,
Filip Byr\'{e}n,
Guillermo Rodas,
Hannes Ingelhag,
Henrik Landgren,
Joar Wandborg,
Love Larsson,
Lucas Magnum,
Niklas Skaar,
Peter Finnman,
Sarah Bernelind,
Olof Hernell,
Pietro Casella,
Richard Stahl,
Sebastian Lindblom,
Sven T\"{o}rnkvist,
Ylva Lundeg\r{a}rd.
Additionally, the first author would also like to thank Xiaolong Liu (Intel Labs) and Xiaoxue Li (Shanghai University of Finance and Economics) for the initial discussion around PU learning methodologies and their applications.

\bibliography{anthology,custom}
\bibliographystyle{acl_natbib}

\end{document}